\documentclass[letterpaper, 10 pt, conference]{ieeeconf}  

\IEEEoverridecommandlockouts                              

\usepackage{graphicx} 
\usepackage{amsmath} 
\usepackage{subfig}
\usepackage{xcolor}

\usepackage{url}

\usepackage{listings}
\usepackage{algorithmic}
\title{\LARGE \bf
Meta-Modeling of Assembly Contingencies and Planning for Repair
}

\author{Priyam Parashar$^*$, Aayush Naik, Jiaming Hu and Henrik I. Christensen
\thanks{The authors would like to thank RPDC and Dr. Nahid Sidki for supporting this research}
\thanks{$^*$University of California - San Diego, 9500 Gilman Dr, La Jolla, CA 92093
        {\tt\small pparasha@ucsd.edu}}%
}

\begin{document}

\maketitle
\thispagestyle{empty}
\pagestyle{empty}

\begin{abstract}

The World Robotics Challenge (2018 \& 2020) was designed to challenge teams to design systems that are easy to adapt to new tasks and to ensure robust operation in a semi-structured environment. We present a layered strategy to transform missions into tasks and actions and provide a set of strategies to address simple and complex failures. We propose a model for characterizing failures using this model and discuss repairs. Simple failures are by far the most common in our WRC system and we also present how we repaired them.

\end{abstract}

\section{Introduction}
Robotics is well known for its performance gains in large scale homogeneous manufacturing. Small scale manufacturing with series of 1-10 is much harder to automate as programming has required too much resources in comparison. The golden standard would be a system where programming time is less than execution time for one item. 
Recently, World Robot Summit (WRS) started an assembly track (called World Robot Challenge, WRC)\footnote{\url{https://worldrobotsummit.org/en/wrs2020/challenge/industrial/assembly.html}} that poses stereotypical examples of small scale manufacturing tasks as the problems to be solved.
We chose to model our system as a series of hierarchical planning, perception and action loops with a reactive behavior tree (BT) framework \cite{colledanchiseBehaviorTrees2018} acting as the executor to catch failures and apply repair.
Our system uses knowledge-based deliberative planners as its backbone with a formal mission definition \cite{melloCorrect1991} and a robotic assembly skills taxonomy \cite{huckabyKnowledge2014} defining the problem and solution for a given assembly product, respectively (Sections \ref{sec:kb}, \ref{sec:planning}).
In order to handle the changing product requirements we devised a generic model of assembly contingencies which can arise with changing product requirements.
We discuss two specific kinds of failure or contingencies in this paper: simple and complex (Section \ref{sec:contingency_recovery}).
We ground the contingency model using a three-tiered architecture we implemented for WRC 2020 (Section \ref{sec:architecture}).
Finally we present our experience with monitoring and accounting for a subset of these contingencies in the WRC 2020 domain as a case study of applying this contingency model to an overly constrained assembly application (Section \ref{sec:wrc_experience}).

\section{Motivating Problem}
\begin{figure}[t]
    \centering
    \subfloat[Initial State\label{fig:wrc_init}]{%
    \framebox{
    \includegraphics[width=0.44\linewidth]{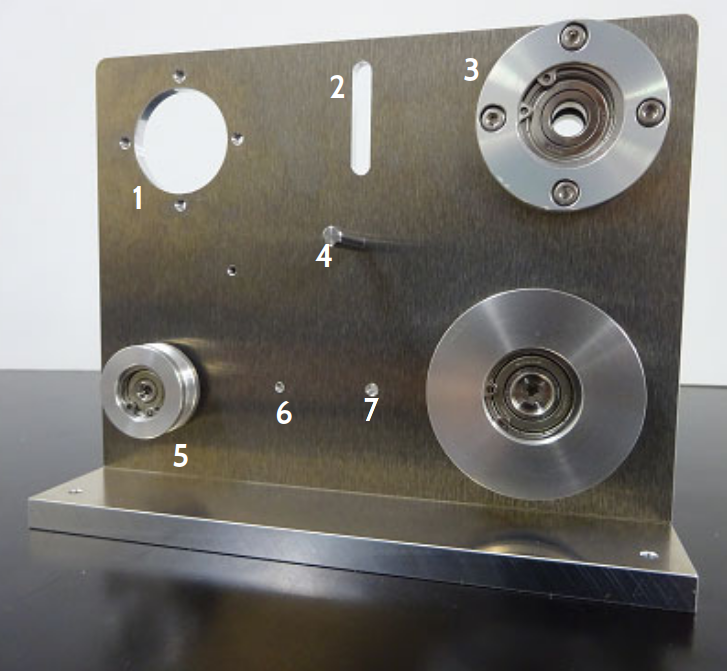}
    }}
    \subfloat[Goal State\label{fig:wrc_goal}]{%
    \framebox{
    \includegraphics[width=0.43\linewidth]{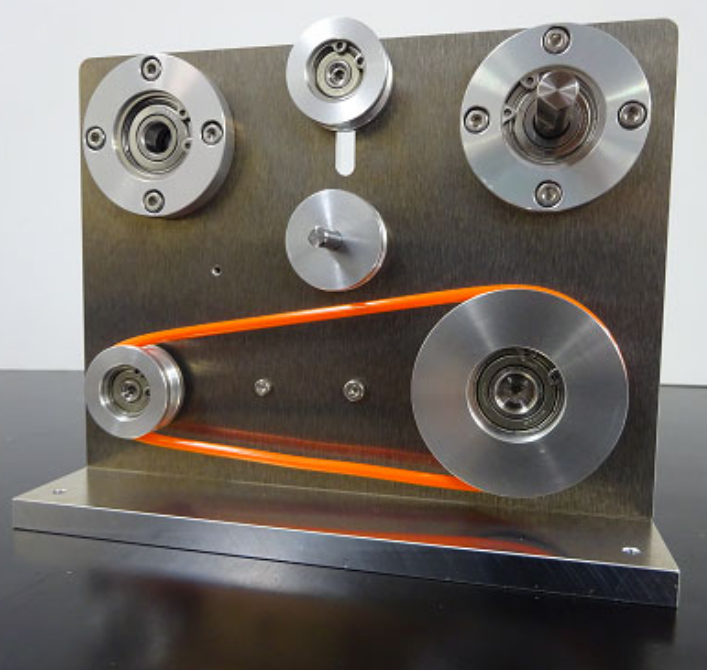}
    }}
    \\
    \subfloat[Housing Asymmetry\label{fig:wrc_housing}]{%
    \framebox{
    \includegraphics[width=0.3\linewidth]{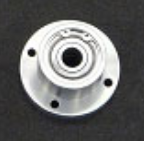}
    }}
    \caption{Example Task. Images from WRC 2020 rulebook \cite{wrs2020rules}}
    \label{fig:wrc_task_board}
\end{figure}

Consider the example WRC task-board assembly problem shown in figure \ref{fig:wrc_task_board}. There are a number of interesting planning problems in this challenge which require: (a) coordination (\#2), (b) part sequencing (\#1), and (c) ability to model abstract tasks (\#5). Further some parts could only be attached to the task-board when picked up in specific configurations, like the housing in Fig. \ref{fig:wrc_housing} which had a wide base and narrow body and only the body was compatible with insertion action.

Homem de Mello presented a graph-based formulation of assembly mission problem in \cite{melloCorrect1991} where the edges encode ``blocking'' relations between parts (e.g. screwing screws into the 4 holes in \#1 before inserting the housing would block stable placement of housing). While this formulation takes care of our constraint (b) it does not cover the rest. If we consider the detailed task-modeling framework in \cite{huckabyModeling2014} we get access to ``coordination'' skills for robots covering constraint (a) but on its own it does not cover all the constraints. We can devise a better assembly framework by combining these two formal representations, but still need to address constraint (c). Further we need object-oriented information for validating and verifying grasped object configuration due to second-order object constraints for attachments as mentioned above. Beyond the planning challenges, the final system would still need to send the task and object constraints down to motion planners and verify events from sensor observations.

\section{Related Work}

\noindent \textit{Assembly planning:} Assembly planning was made tractable using knowledge-based representations like petri nets \cite{zhangRepresentation1989} and its successor AND/OR \cite{melloRepresentations1991} graphs which efficiently represent the events, constraints and transitions in assembly sequences. These representations made the search for legal orderings of an assembly goal-sequence faster. Later to afford greater control over the methods of actuating assembly tasks, these high-level planners were often integrated with classical planning \cite{bylanderComputational1994, nau1999shop} for automatic generation of assembly programs \cite{huckabyModeling2014, dantam2012linguistic}. Assembly product can be defined as the final goal state but getting to it requires an understanding of not only inter-part relations but also geometries and the attachments and contacts that they enable. Thus two main representations: constructive solid geometry (CSG) \cite{hoffman1989automated} and boundary representation (B-Rep) \cite{hoffman1990automated} were introduced to integrate task knowledge with object-knowledge. While very powerful, especially since they allow inferring possible tasks related to object from mere descriptions \cite{perzyloOntology2015}, both of these representations require detailed geometric modeling ahead of time to be used for automatic assembly planning. One of the constraints for WRC surprise challenge was that teams would know dimensions of parts ahead of time but not detailed models, thus motivating representations beyond the traditional ones with maybe less knowledge but more flexibility. \cite{lallement2019combining} present a good review of challenges in communicating high-level constraints to metric planners. We use a simple method to forward constraints but use reactivity of the behavior trees to track the consistency of metric-level detail with high-level task expectations and repair plan if necessary. We use external functions like \cite{dornhegeIntegratingTaskMotion2010} and inter-weave them in the behavior tree representation of primitive actions as pre and postconditions.


\noindent \textit{Assembly Contingencies \& Recovery:} \cite{heger2010robust} uses mobile robots and repairs for contingencies specifically arising from failed motion planning due to necessary simplification of the planning problem. Our meta-model of contingencies takes a complementary view of failures where they arise due to transformations of the planning problem. We also present a preliminary rule-book on when does a failure need to be moved up or down the assembly planning ladder. \cite{kootballyRobust2015} presents how task failures can be enumerated along with reasons and remedies for a kitting task and this knowledge can be used for repairs during execution-time. While a kitting task does not have second-order relations of typical assembly problems (e.g. object-configuration in hand affecting success of attachment task) this is an example of how lower-level failures can be lifted for planning resolutions.

\section{System Architecture}
\label{sec:architecture}

\begin{figure}[t]
    \vspace{0.1in}
    \centering
    \framebox{
    \includegraphics[width=0.4\textwidth]{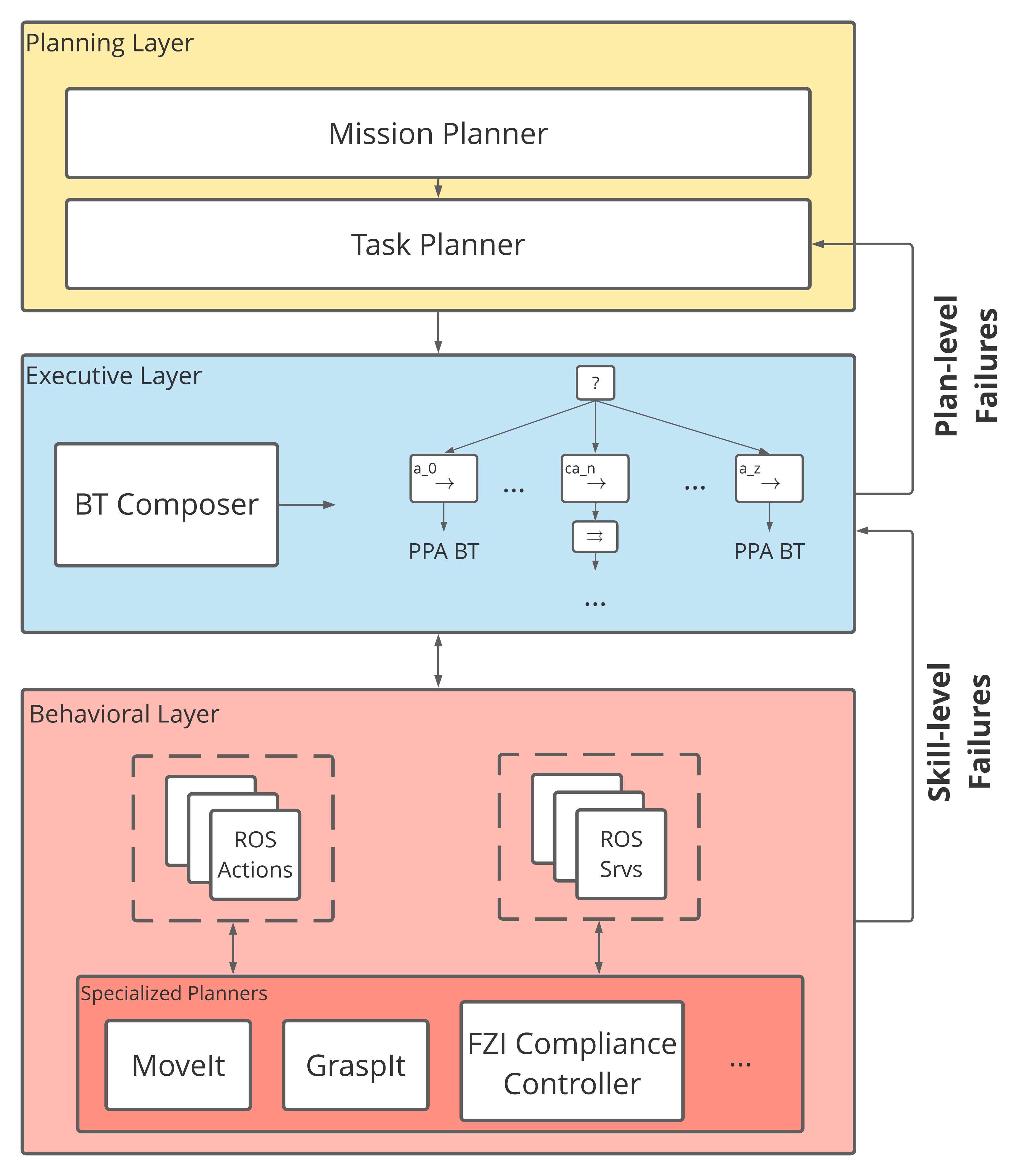}
    }
    \caption{Hierarchical system architecture showing the organization of planning components with their nominal dependence on each other (arrows down the center) and the arrows on right showing their dependence when a failure or contingency occurs}
    \label{fig:wrs_sys_arch}
\end{figure}
\begin{figure*}[t]
    \vspace{0.1in}
    \centering
    \framebox{
        \includegraphics[width=0.92\textwidth]{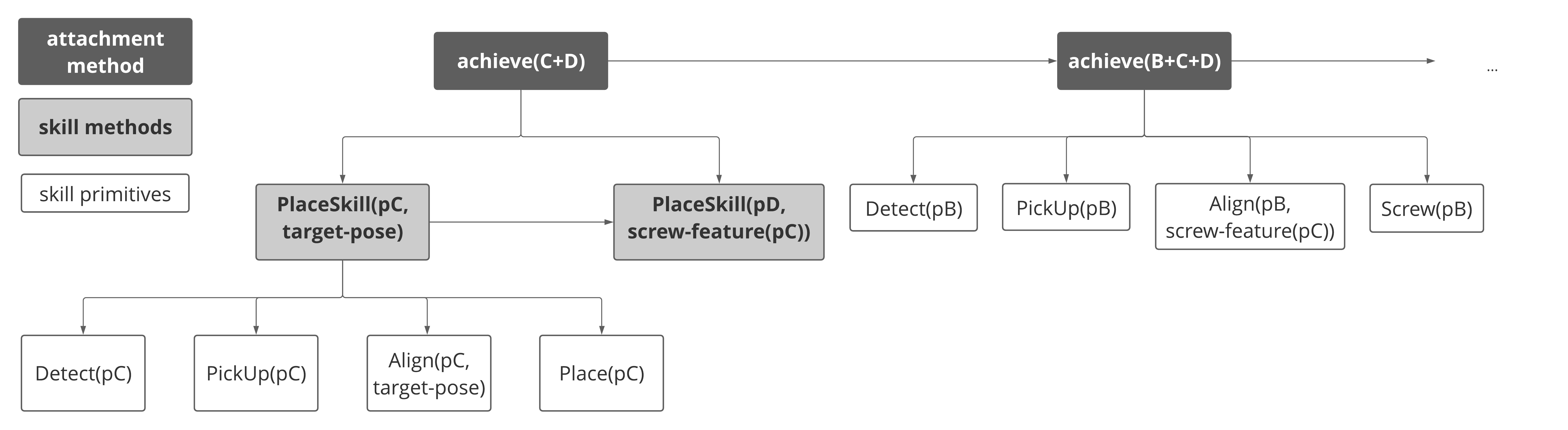}
    }
    \caption{Task plan for placing C on D for Mission in \ref{code:mission_descr}}
    \label{fig:mot_task_plan}
\end{figure*}

Closely based on the three-tiered architecture of robotic planning systems\cite{kortenkamp2016robotic}, each stage has its own database of knowledge and data that it uses, the output of previous informing the subset of possible choices for the next stage, and observations at the lower rungs dictating decisions made at higher-levels.
Our planning layer is a to component composite, the mission planner consumes a description of the assembly problem and does long-term planning by ordering part placements with respect to each other. The mission ordering is then described as a collection of sequential tasks by the task planner. Each task is made of assembly skills that the robot has been programmed for.
The executive layer converts the skill-level plan to behavior trees (BTs)\footnote{\url{https://github.com/BehaviorTree/BehaviorTree.ROS},\\ \url{https://github.com/BehaviorTree/BehaviorTree.CPP}} which interface with specialized planners executing the intended robotic primitive actions for the assembly skill.
Finally the behavioral layer is where the nodes of the BT or robotic primitive actions are implemented in software using ROS. This level encodes everything which is required to convert task-space commands to joint-space commands and sensor observations to higher-levels assertions about environment-state conditions.

\subsection{An Example Walk-through}
\begin{figure}[b]
\centering
\framebox{
    \includegraphics[width=0.5\linewidth]{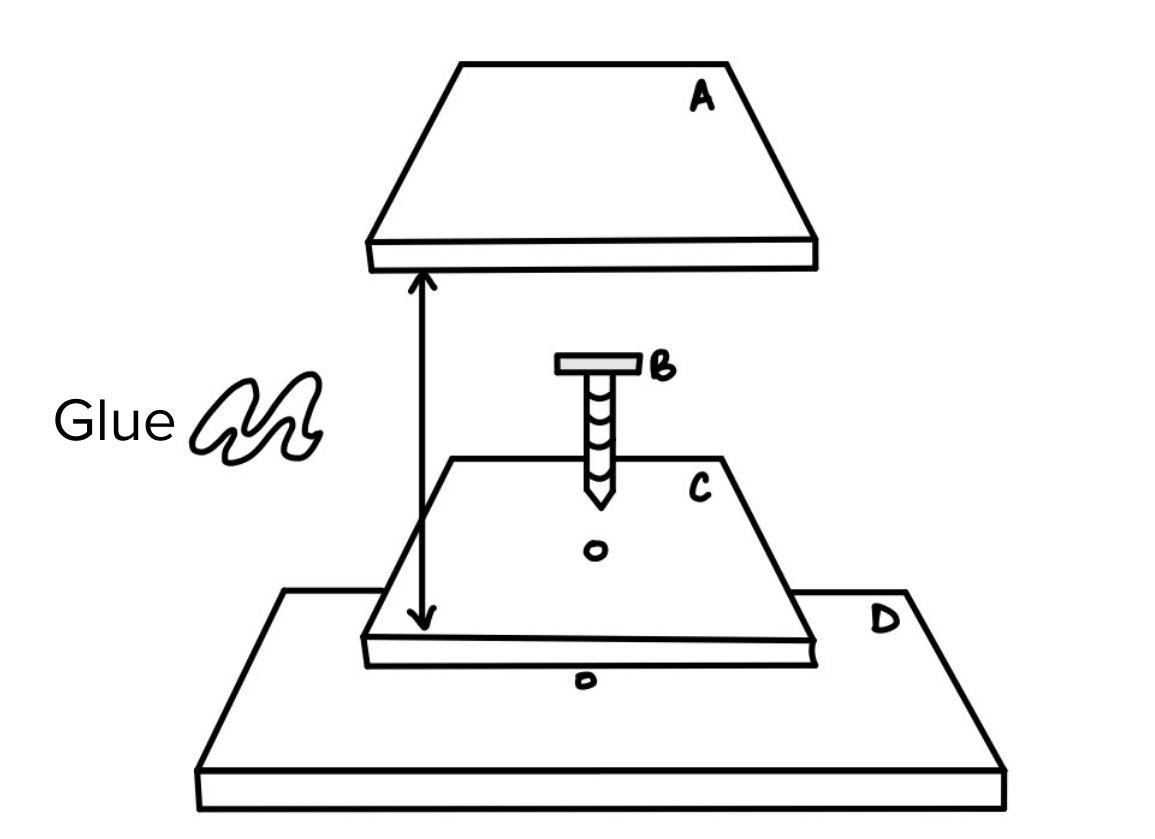}
}
\caption{Motivating Assembly Problem}
\label{fig:mot_assembly_problem}
\end{figure}
Consider the assembly problem in Fig. \ref{fig:mot_assembly_problem}. The mission planner is provided information about the product as follows (we only present the information relevant to the walk-through):
\begin{lstlisting}[caption={Example Mission Description}, captionpos=b, label=code:mission_descr]
 parts:
   pA: plain-tile
   pB: screw0
   pC: tile-with-hole
   pD: tile-with-hole
 attachments:
   a0: pA, pC
   a1: pB, pC
   a2: pB, pD
relations:
  blocked-by:
    a1: pA
    a2: pA
\end{lstlisting}
 
The mission planner ingests this information and returns an ordering:

\begin{center}
    \texttt{C+D $\rightarrow$ B+C+D $\rightarrow$ A+B+C+D}
\end{center}

To assemble \texttt{C} onto \texttt{D}, we would need a task plan enumerating all the steps for picking, placing and alignment as shown in Fig. \ref{fig:mot_task_plan}.

\begin{figure}[h]
    \centering
    \framebox{
        \includegraphics[width=0.6\linewidth]{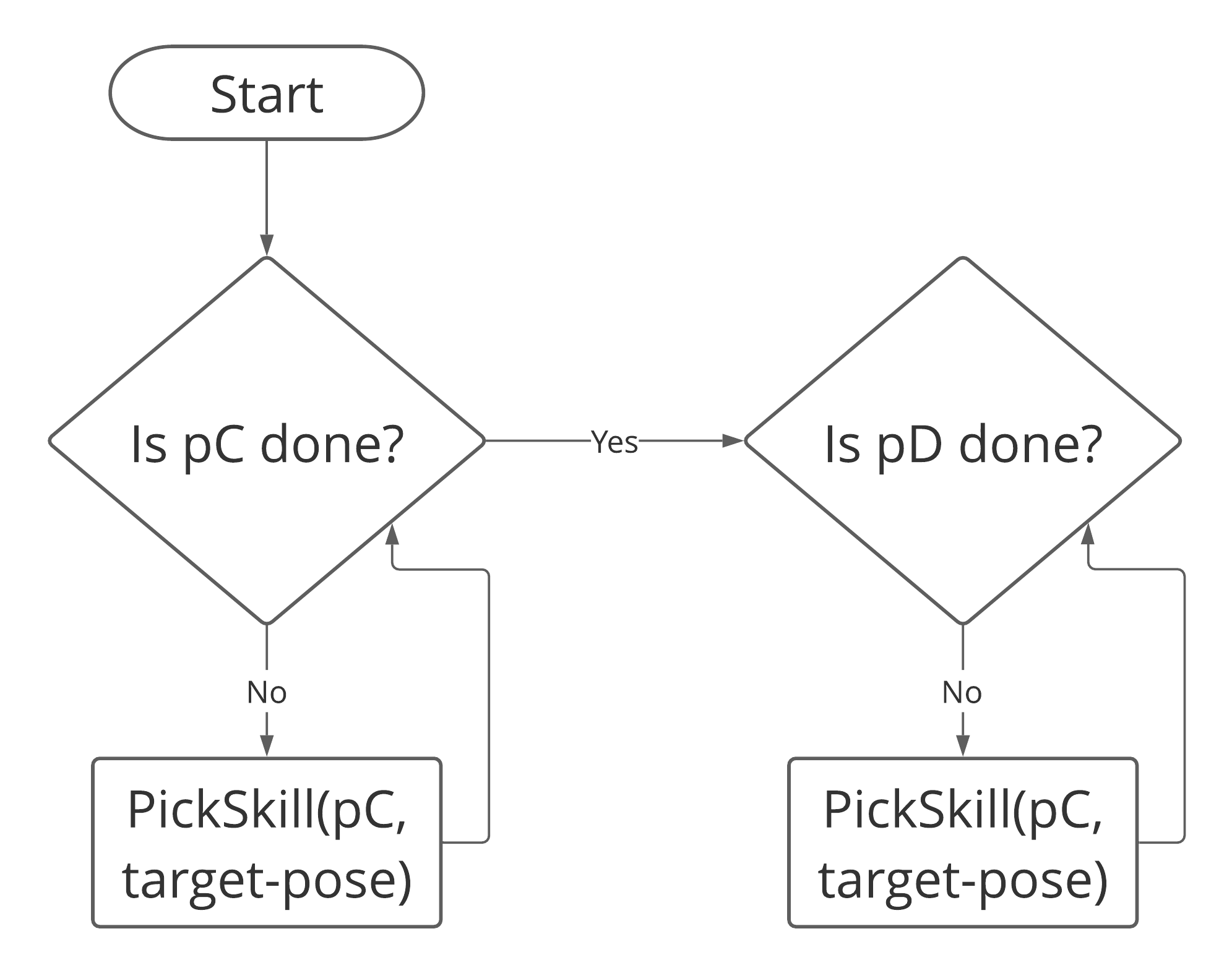}
    }
    \caption{BT representation of task-plan in Fig. \ref{fig:mot_task_plan}}
    \label{fig:mot_bt}
\end{figure}

Finally, this is instantiated as a BT which includes injection of physical parameters (e.g. the exact pose of \texttt{screw-feature(pC)}) and special condition nodes to verify progression of the task-plan. A depiction based on the motivating example is shown in Fig. \ref{fig:mot_bt}. We use a decision-tree like syntax in this diagram to avoid explanation of BT semantics.

The decision nodes in Fig. \ref{fig:mot_bt} keep invoking \texttt{PickSkill} until the ``done'' condition of \texttt{pC} is true. A BT node returns three states: \textit{running}, \textit{success} and \textit{failure}, so the constant polling of \texttt{PickSkill} enables us to monitor plan progression on an arbitrarily fine time-scale. This is an important property of BT which we will exploit for reactive recoveries from assembly execution failures later. In the next two sections, we go over the each of the knowledge-representations shown here and formalize the knowledge-bases and planning processes supporting this walk-through. While this example shows the process for only one leg of the plan, the process is repeated for every mission event and converted into one large BT.

\section{Knowledge Representations}
\label{sec:kb}
Going from the lowest to the highest level of abstraction, this section describes the knowledge-bases and problem specifications for each of them.

\subsection{Primitive Actions of Assembly Domain}
We use the taxonomy proposed in \cite{huckaby2012taxonomic} to describe a single agent's assembly action-space. The taxonomy provides guarantees on completeness and soundness on the Cranfield Benchmark \cite{huckabyModeling2014, collins1985development}, making it an excellent candidate for our domain. The taxonomy defines general assembly skills and decomposes each skill into specific skill primitives required to actuate that skill in different contexts. The assembly skills were extended to suit our purposes (like adding mounting task) and organized as following (in order of skills in a task):
\begin{enumerate}
    \item Perception/Detect
    \begin{itemize}
        \item Object
        \item Force
        \item Grasp
    \end{itemize}
    \item Pick-Up(object)
    \begin{itemize}
        \item Compute-Grasp
        \item Grasp/Close grippers
    \end{itemize}
    \item Move
    \begin{itemize}
        \item Transport
        \item Move until force feedback $>$ threshold
        \item Servo based on visual feedback of object/feature
    \end{itemize}
    \item Align (with respect to object or feature)
    \item Fasten
    \begin{itemize}
        \item Transition-fit Insert
        \item Screw
        \item Mount (elastic)
    \end{itemize}
    \item Release-object/Open-grippers
    \item Verify task
    \begin{itemize}
        \item Pre-conditions
        \item Post-conditions
    \end{itemize}
\end{enumerate}

\subsection{Task Representation: HTNs and BTs}
\label{sec:repr_pa}
Each skill listed in the previous section is described as a high-level method using HTN terminology \cite{erol1995semantics} and decomposes into skill primitives with correct argument based on domain preconditions. Each skill primitive is formally defined as:
 $$
 pa = (head(pa), pre(pa), post(pa)),
 $$where $head$ is the name, and $pre()$ and $post()$ are function mappings to preconditions and postconditions which are in the conjunctive normal form (CNF). The ``is \texttt{pC} done'' condition is a post-condition of PickupSkill in Fig. \ref{fig:mot_bt}. Each skill is then formulated as a simplified method defined formally as:
 \begin{align*}
     skill_m = (head(skill_m), pre(skill_m),\\ pa(skill_m), post(skill_m))
 \end{align*}
where $head$ is same as in skill primitive definition, and the $pre(skill_m), pa(skill_m), post(skill_m)$ together define the rules when one skill primitive is preferred over others for the same skill and what is the expected result of execution. We have added the $post(skill_m)$ as an extension to existent HTN semantics to make the plan-execution more closed-loop and support verification after every high-level action to account for uncertainties. At the top of this hierarchy are attachment-methods which hierarchically decompose to find the valid plan for the mission leg. Similar to \textit{goal-methods} in HTN semantics, this method is read as \textit{achieve(attachment-symbol)} making completion of provided attachment a necessary post-condition for plan progression. In the following excerpts, line numbers 2, 4 and 6 are examples of pre-conditions specializing skill to a specific primitive.

\begin{lstlisting}[language=Python]
def fasten(part, agent):
  if is_screw(part) or is_bolt(part) or is_nut(part):
    screw(part, agent)
  elif is_housing(part) or is_shaft(part) or is_pulley(part):
    insert(part, agent)
  elif is_elastic(part):
    mount(part, agent)
  return False
    
skills[`fasten'] = fasten
\end{lstlisting}

Line number 2 is a precondition while 2 and 3 are examples of post-conditions expected to render true after successful execution of \texttt{screw}.
\begin{lstlisting}[language=Python]
def screw(part, agent):
  if in_hand(part, agent) and is_aligned(part):
    part.state = screwed
    agent.state = done
\end{lstlisting}

Specialized predicates are used in these conditions which enable the planner to query the state of object and agent. 
\begin{lstlisting}[language=Python]
def is_screw(part):
  if part.name == 'screw':
    return True
  return False
\end{lstlisting}

We used BTs to instantiate plans at run-time due to two major reasons:
\begin{enumerate}
    \item Decisions deferred at planning time could be represented using condition nodes and complex control-flows at run-time
    \item Ability to pre-compile reactive control-flows for local recoveries 
\end{enumerate}
We used the methodology described in \cite{colledanchiseBlendedReactive2019} to translate each descriptive action into a Postcondition-Precondition-Action (PPA) BT. So each primitive action was stored as a reusable sub-tree in plan library; we discuss how this enabled us to do verification in Section \ref{sec:htn2bt}.
 

\begin{figure*}
    \vspace{0.1in}
    \centering
    \includegraphics[width=0.95\textwidth]{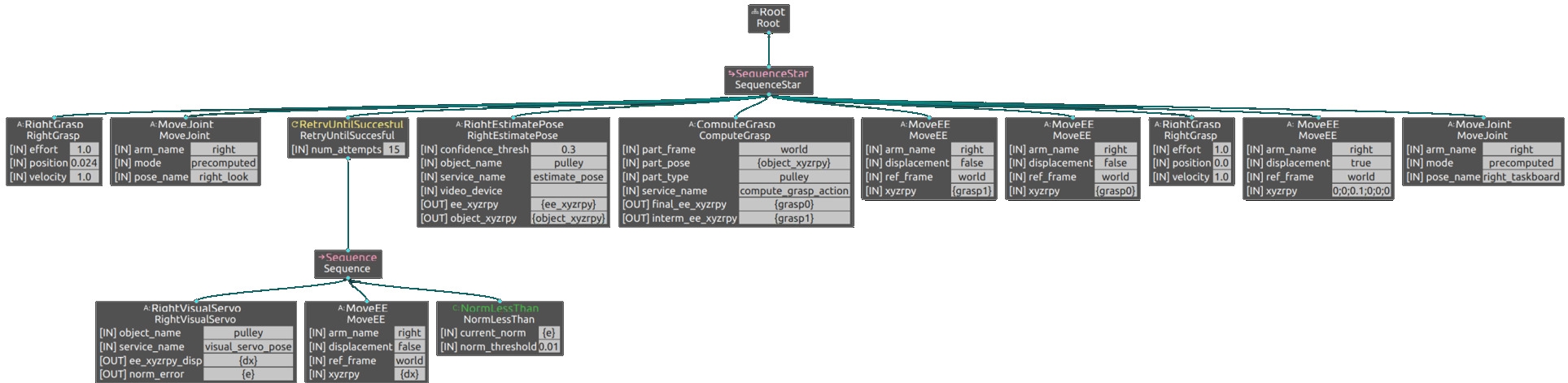}
    \caption{Actual Behavior Tree Expansion for the Pick-Up Skill}
    \label{fig:bt_example}
\end{figure*}

\subsection{Mission Representation of the Product}
\label{sec:mission_repr}
Based on \cite{melloCorrect1991} we devise a refined formalism for formulating a dual-arm assembly mission planning problem. Such a representation is one-level above the semantic descriptions in \cite{perzyloIntuitive2016} which define detailed physical aspects of parts and related geometric constraints. All the parts involved in the assembly product are collected in a set $P$, these parts can be simple or complex in nature, i.e. arbitrarily large or small in terms of number of features. Every interaction required between parts to build the product is called an \textit{attachment}, all the attachments required for the product are collected as a set $A$. For each attachment we define following attributes:
\begin{LaTeXdescription}
    \item[Type of attachment] The structural type of interaction between parts: \textit{screwing, cylindrical insertion, mounting}
    \item[Part Relation] List of parts involved in attachment
    \item[Blocked-by Relation] If an attachment $a_n$ is associated with part $p_x$ by this relation then executing any attachment related to this part will $a_n$ impossible in future
    \item[Stability-dependence Relation] A new attribute introduced to formulate the three-part planning required by the WRC task (\#2 in Fig. \ref{fig:wrc_goal}); it means the associated parts are required to be attached (almost) concurrently
\end{LaTeXdescription}
The above mentioned relations of an attachment are collected as named-sets in set $R$. 


\section{Planning}
\label{sec:planning}
\subsection{Mission Planning}
Given the modeling in previous section, a partial-order planner uses the \textit{blocked-by} and the \textit{stability-dependence} relations to establish causality links between the attachments. Each attachment of the assembly product is added as a node to a planning-graph. If an attachment and a part are associated using \textit{blocked-by} relation then a \textit{must-precede} causal link is set up between the given attachment and all the attachments that the blocking part implicates. Similarly, if two attachments are related by \textit{stability-dependence} relation they are clustered into one \textit{complex attachment} enforcing complete execution of the sub-attachments before control moves on to next attachments in the plan.
Once attachments are causally linked following this scheme we topologically sort the graph and output a least-commitment plan.

\subsection{Task Planning}
We use a python version of the SHOP planner for this component\footnote{\url{https://bitbucket.org/dananau/pyhop}}. Newer versions of SHOP relax total-ordering constraints and have more expressivity, but the benefit of being able to use Python for rapid prototyping of plans and lack of high-level complexity in WRC tasks made PyHOP sufficient for our needs. Initial state of the assembly is depicted in a Python data-structure with mission description and partial-order from the level above already filled out.
Since our system is dual-armed we also need to perform resource arbitration between two agents. This is done using a simple heuristic: any task approaching from the front of task-plane is done by the right arm and left arm is allocated to everything else.
Bulk of task planning is done by the top-level method \textit{assembly-plan} which chooses the next not-done, conflict-free attachment based on mission ordering and iteratively builds the final HTN.

\subsection{Behavior Trees, Verification and System Processes}
\label{sec:htn2bt}
Certain planning decisions can not be taken until observations are made in the real-world, so how do we instantiate plans? We manually annotate decision-points in HTN domain which require perception and use backchaining algorithm, as described in \cite{colledanchiseBehaviorTrees2018}, to create an \textit{if-then-else} equivalent of BTs based on different pre-conditions of the decision point. Each BT node interacts with specialized planners, vision routines and actuators using ROS middleware, describing this layer is out of scope of the current paper.
Figure \ref{fig:bt_example} shows the nominal behavior tree for pick-up skill before it is expanded to account for contingencies.

\begin{figure}
    \centering
    \framebox{
    \includegraphics[width=0.8\linewidth]{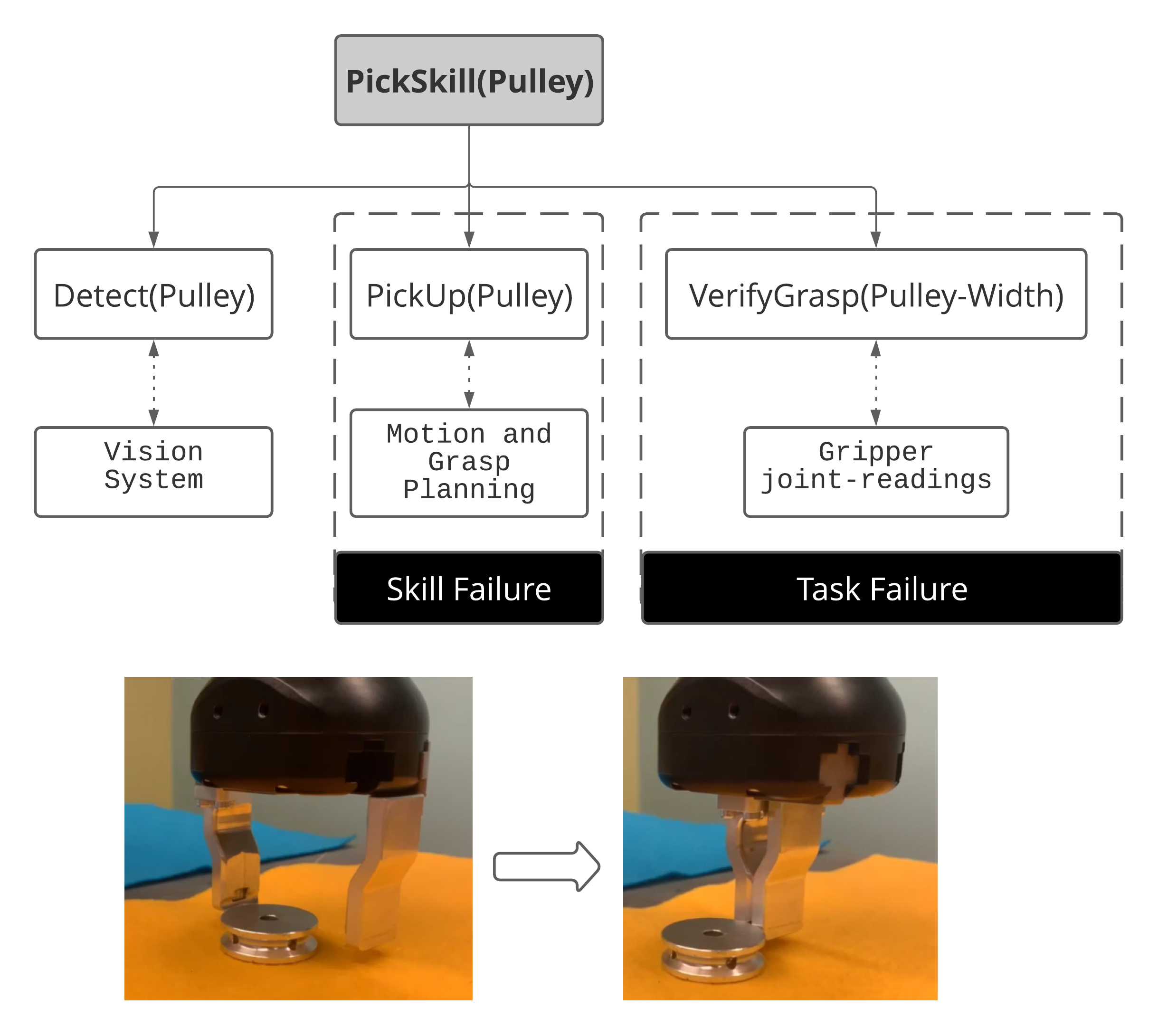}
    }
    \caption{A description of how BT monitors for skill and task failures. If the motion or grasp planner fail the failure is propagated up from the action node (PickUp) however if the failure is due to object not being grasped, as depicted in images below the BT, then failure propagates from condition node (VerifyGrasp) failure.}
    \label{fig:verification_bt}
\end{figure}

To monitor and verify the plan progression we use two kinds of signal-checking methods in our BTs. Every \textit{action node} which communicates with a lower-level planner, lifts its errors as a \textit{failure} signal in the BT. This allows us to verify the actuation of planned actions. Next, we implemented the pre and postconditions of the skills as \textit{condition nodes} which assess the output of an action node against expected value for skill execution/completion. This method allows us to verify that actuation occurred but also changed the environment in the intended way. Figure \ref{fig:verification_bt} shows an example of this for \textit{PickUp} skill with an example of actuation success leading to task failure. Another verification primitive \textit{VerifyPose} helped in verifying that parts were in correct configuration (like Fig. \ref{fig:wrc_housing}) for the task.

\section{Modeling of Contingencies and Recovery}
\label{sec:contingency_recovery}

Given the different functions of planning layers, how may we define one model of contingencies which applies to all three of them? We choose to do this by first defining a meta-model over these representations which characterizes the space of domain knowledge and domain elements allowing planning and execution to be done. Then we define contingencies over this meta-model grounded in specific examples. Our meta-model complements the MacGyver problem formulation put forth by \cite{sarathy2018macgyver}, which focuses on how to repair plans, by defining the situated space of failures in an assembly planning problem. 

Our meta-model of assembly planning representations consists of three orthogonal axes:
\begin{itemize}
    \item \textit{Procedural-space:} The knowledge about how to \textit{do} mission sequencing or task procedure or actuation of skill primitive.
    \item \textit{Elemental-space:} The knowledge relating procedures with the features of and relations between domain elements or objects.
    \item \textit{Grounding-space:} This is where we take the general knowledge from procedural and elemental space but then \textit{ground} it in the specifics of the current assembly problem.
\end{itemize}

An ideal assembly agent would have the procedural knowledge about all products, tasks and skills possible, which is of course not realistic nor the goal. A more realistic goal for agile manufacturing is that agents have enough knowledge to recover when the product design, the underlying task shape or the skills change in a small way.


We will introduce some minimal terminology here to keep our subsequent discussions clear. Assume the system plans with incomplete information and reaches a contingent state not consistent with the optimal policy $\pi$ that leads to goal state. This contingency requires some additional knowledge to either replan and get back to $\pi$ or plan for the next most optimal policy, say $\hat{\pi}$, from contingent state to the goal-state.
This section details the symptoms, effects and resolutions for possible contingencies at the three levels of planning when acting with incomplete knowledge and environmental uncertainty. Based on whether the contingency allows transition back to $\pi$ or not we define two modes of contingencies to consider:
\begin{enumerate}
    \item \textit{First-order}: Given additional knowledge (for any of the three axes) a first-order contingency can be reversed to allow the system to continue with $\pi$.
    \item \textit{Higher-order}: No additional knowledge exists which can enable system to go back to $\pi$. Instead the additional knowledge is used to find $\hat{\pi}$.
\end{enumerate}


\section{Hierarchical Grounding of Contingencies for Our Architecture}

\subsection{Motivating Problem}
Consider the problem in Fig. \ref{fig:mot_assembly_problem}. In this section we introduce first and higher-order contingencies in the planning for this problem and discuss the recoveries.
The current methods and paradigms of doing assembly place heavy emphasis on correctness of lower-level details for overall plan to succeed. In order to relax these assumptions it is necessary to characterize the nature of contingencies which can be expected.
Our meta-model provides the language so that contingencies and recoveries can be characterized in precise terms and given these axes of characteristics we can more systematically analyze the generality of a given system or methodology.
\subsection{Mission-level Contingencies}

Primary knowledge source exploited at the mission-level is the structure of planning graph as defined by \textit{blocked-by} relation.
Let's assume the \textit{grounding} for this parameter is missing and consider the contingencies which can occur while planning for the given example.
Please note that the contingency is found during the execution time even though the failure occurred during planning phase.
\begin{enumerate}
    \item First-order: The agent screws \texttt{B} onto \texttt{D} making \texttt{B+C+D} physically impossible. A repair would require task knowledge about effects of \textit{reversing} tasks to plan a mission where we unscrew part \texttt{B} and restore environment back to a state consistent with $\pi$. Thus, we require \textit{complementary knowledge in the procedural-space} to reverse or undo the actions done. However the elemental and grounding knowledge does not transform because of this error, meaning the physical properties of \texttt{B} and its effects on helical screwing remain the same and the pose grounding of unscrewed \texttt{B} with respect to \texttt{C} is also the same. 
    \item Higher-order: Agent glues \texttt{A} onto \texttt{C}, later realizing this makes $\pi$ impossible due to the irreversibility of this attachment. In this situation the remaining mission has an optimal ordering $\hat{\pi}$ that is strictly sub-optimal for the original mission. A possible sub-optimal plan would look like:
    
    \begin{center}
        \texttt{A+C $\rightarrow$ A+C+D $\rightarrow$ A+C+D+B},
    \end{center}
    This recovery requires changing the \textit{grounding} of the screw skill primitive since the expected direction of action is blocked off, however the applicability of screwing procedure is consistent with respect to objects.
\end{enumerate}


\subsection{Task-level Contingencies}

\begin{figure}[ht]
    \centering
    \includegraphics[width=0.75\linewidth]{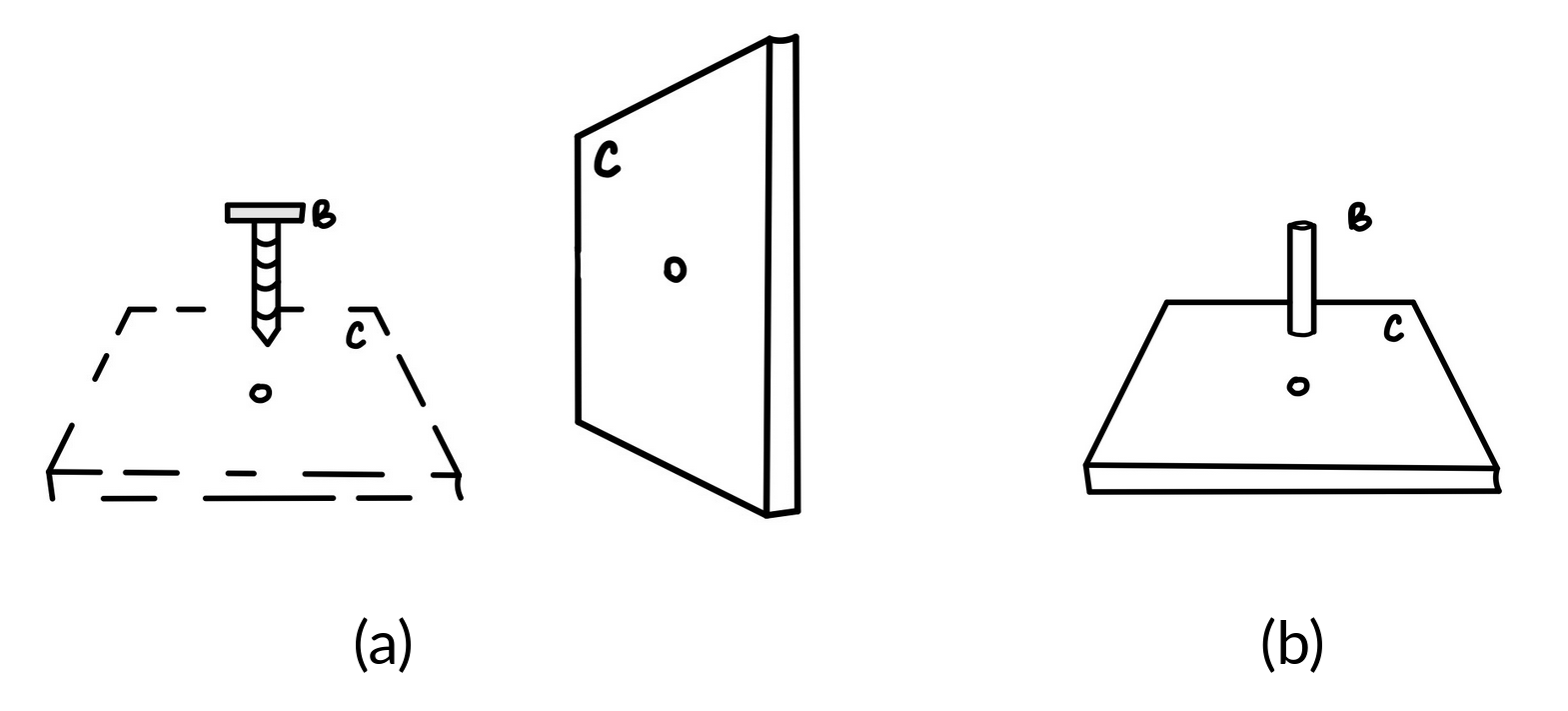}
    \caption{Examples of first and higher order task contingency}
    \label{fig:task_contingency}
\end{figure}

We consider two simplification assumptions made by task planners to make an assembly problem tractable: positions of objects and availability of specific resources. 

In our running example, consider the task procedure for \texttt{B + C}. The planner decomposes this through \textit{Screw} task where the task-model makes an assumption that \texttt{C} would be horizontally aligned to simplify the plan. However when the agent picks up \texttt{B} and reaches \texttt{C} agent realizes that \texttt{C} is actually vertically aligned (figure \ref{fig:task_contingency}(a)). This is an example of a first-order contingency where the agent can repair part placement (\textit{grounding}) and resume $\pi$.
More generally, the correct task's pre-conditions are not satisfied in the environment, so the agent defines a sub-goal of enabling the pre-condition as an intermediate step before resuming with the nominal plan.

Consider another assumption made by the planner where it assumes that the \texttt{B} will necessarily be a screw, however an updated ensemble has it replaced with a dowel (figure \ref{fig:task_contingency}(b)). This is a simple replacement but significantly affects planning with respect to expected $\pi$. If 'dowel' exists in the elemental knowledge-base of the planner then a simple task replan with updated part description would find the new $\hat{\pi}$. However, if dowel is not in the preconditions of \textit{Insert} skill primitive then the agent would need advanced repair where it infers applicability of \textit{cylindrical insertion} attachment from the geometry of a dowel.

\subsection{Skill-level Contingencies}

\begin{figure}[hb]
    \centering
    \includegraphics[width=0.75\linewidth]{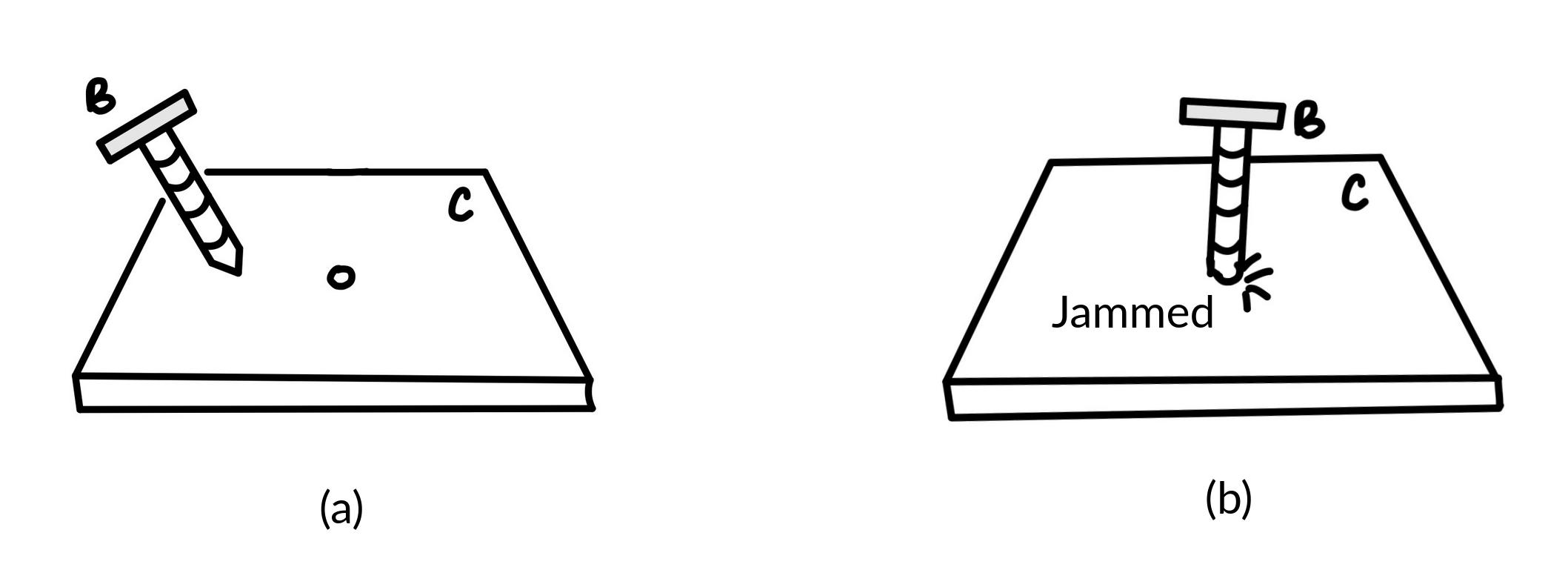}
    \caption{Example of first and higher order skill contingency}
    \label{fig:skill_contingency}
\end{figure}

A first-order skill-level contingency is marked by the ability to reset the action without manual intervention. An example is when agent is trying to align screw \texttt{B} with the threaded hole in \texttt{C} but fails so the screw drops out (figure \ref{fig:skill_contingency}(a)). Any state where the policy of the skill can be \textit{re-grounded} by simple transformations of 6D poses (\textit{PickUp} from $pose_{1}$ instead of $pose_{0}$) and be expected to work falls under a first-order contingency. Such recoveries promise probabilistic completeness in the absence of extensive dynamics modeling.

A higher-order contingency is where the procedural or elemental spaces need to be expanded such that the available parameters can include information about new \textit{physically meaningful} entities. To elaborate, consider a novice agent using rigid-controls to screw \texttt{B} and due to limited control the screw gets jammed within the threaded hole (figure \ref{fig:skill_contingency}(b)). In order to repair this action such that the contingency does not happen again the procedure-space of controls for screwing would need to be reformulated to consider \textit{compliance} as an additional parameter. This is a very hard problem and tests the extremes of creative repairs and planning in robotics literature.


in

\section{A Case Study: World Robot Summit Assembly Challenge}
\label{sec:wrc_experience}
The WRC's problem statement is to push for technology which satisfies industry 4.0 requirements \cite{liao2017past} specifically one of flexible programming with accurate execution for low-tolerance assembly problems.
This particular application was of an over-constrained assembly problem where the system was expected to finish most number of tasks with accuracy, and repeatability, in least amount of time possible.
Our system was successfully able to instantiate plans for the task-board challenge shown in Fig. \ref{fig:wrc_goal} and was selected as one of the 16 finalists based on our robust performance.
To optimize run-time our contingency hierarchy was used in this system as a way of keeping the system stable and running in non-fatal, non-contingency regions with majority of the symptom monitoring, recovery knowledge and plans pre-compiled into its planning stack.



\subsection{Assumptions and Contingency Scope}

We assume complete knowledge about the design parameters of the product, which is a reasonable assumption given the lack of complex dependence between parts in the challenge mission (figure \ref{fig:wrc_goal}). The competition is also very specific about what kind of part interactions it wants to test for example screwing and cylindrical insertion amongst others. So we assume our system will always operate within the knowledge-space of the behaviors at the lowest-level. Due to these assumptions our system was most susceptible to first-order task and skill-level contingencies. Table \ref{table:wrs_contingencies} enumerates all contingencies we observed and developed our system to handle. Contingencies were classified following the monitoring rules outlined in Section \ref{sec:htn2bt}. Action failure suggests skill contingency while condition failure suggests task contingency.

\begin{table}[t!]
\vspace{0.1in}
\centering
\begin{tabular}{l l}
    \hline
    \textbf{Contingency Class} & \textbf{Specific Type Observed}\\
    \hline \hline
    First-order Task & 1. Parts not placed in right configuration\\
                     & 2. Part-stations not at expected location\\
                     & 3. Part not in hand after grasping for next task\\
                     & 4. Alignment action resulted in failure\\
    \hline
    First-order Skill   &   1. Object not detected due to lighting\\
                        &   2. Failed grasp actuation\\
    \hline
    Unclassified        &   1. Hardware-drivers dropping out\\
                        &   2. Sensor noise due to environmental frequencies\\
    \linebreak
\end{tabular}
\caption{List of classified contingencies as observed in WRS system}
\label{table:wrs_contingencies}
\end{table}

\subsection{Recovering from Contingencies}

In theory, once the task-level contingencies are pushed up to the task-planner, it replans a new high-level plan which is then translated and stitched back into the behavior tree using backchaining \cite{colledanchiseBlendedReactive2019}. However since the developed system was for a competition we did this backchaining ahead of time by enumerating through all possible configurations of part placement possible and instantiating a complete behavior tree (Section \ref{sec:htn2bt}.

For the skill-level contingencies, we observed majority of the repairs could be made by using special \textit{decorator nodes} in behavior trees which allow one to arbitrarily modify the control-flow, for example, \textit{RunUntilSuccessful} or \textit{RunNTimes}. For the first-order skills listed in table \ref{table:wrs_contingencies} we created reactive strategies which changed the environment by a small amount and retried the same action and observed a marked decrease in fatal failures.
For the listed contingencies following recoveries were seen to work:
\begin{enumerate}
    \item Change the camera position with orientation still aimed at part-station coordinates and retry action
    \item Execute action N times and choose least-error detection
    \item Reset underlying ROS drivers and try again
\end{enumerate}

However, not all task contingencies required deliberation like the latter two listed in table\ref{table:wrs_contingencies}. These are actually task-level representations of contingencies which occurred due to noisy execution or wrongly grounded execution. A ``part not being in hand'' is a task-level characterization of a failed servoing which leads to Fig. \ref{fig:verification_bt}. Thus, when skill-level failures exert a non-local effect they \textit{can become} a task-level contingency with reactive repairs. For example, adding a \textit{RetryUntilSuccessful} decorator over the entire \textit{PickUp} plan segment is a simple way of reactively repairing task effects of compounded skill-level failures. The trade-off is between run-time and computational (sophisticated software) or system (sophisticated hardware) complexity. 

\section{Conclusion and Practical Recommendations}
In this paper we presented our assembly planning system which combines formal representations with behavior trees to instantiate reactive plans based on deliberative knowledge. We also propose a model for characterizing contingencies in assembly within the context of small-lot manufacturing which faces changing problem requirements. Our system is motivated and applied on WRC 2020 challenge, and we present a discussion of observed contingencies and specific recoveries that worked.

Failures at the skill-level can have high costs, especially when actuating force-based contacts, thus we recommend precompiled repair knowledge at this level. Not only does this reduce overall recovery cost of an assembly system but also enables the system to exist in a non-fatal state enabling other sophisticated recoveries like inference and deliberative reasoning to take place. Further, based on our meta-modeling we believe a robust elemental knowledge-base can support both higher-level inferences of task-knowledge and grounding of policies in physical parameters of domain objects. For future work, we want to focus on the details of how to represent and learn elemental-knowledge from geometries, without extensive annotations, such that it can enable task-model inferences. We observed in this system that reusable repair units of BTs enabled significant savings on development and verification time while prototyping new behaviors, and in a similar vein we believe by inferring task-models from instance-based knowledge we can approach more generic plan repairs.


\addtolength{\textheight}{-12cm}   









\bibliography{IEEEabrv, bibliography.bib}
\bibliographystyle{IEEEtran}

\end{document}